\documentclass{vgtc}                          % final (conference style)
\graphicspath{{figures/}{pictures/}{images/}{./}} % where to search for the images

\usepackage{times}                     % we use Times as the main font
\usepackage{tabu}                      % only used for the table example
\usepackage{booktabs}                  % only used for the table example
\usepackage{lipsum}                    % used to generate placeholder text
\usepackage{mwe}                       % used to generate placeholder figuresㅁ
\usepackage{kotex}

\usepackage{amsmath}
\usepackage{amssymb}

\usepackage[dvipsnames, svgnames]{xcolor}
\usepackage{enumitem}
\usepackage{caption}

\renewcommand{\paragraph}[1]{\vspace{4pt}\noindent\textbf{#1.}}

\definecolor{applepinknormal}{RGB}{255, 55, 95}
\newcommand{\revise}[1]{#1}

\onlineid{1049}

\vgtccategory{Research}

\vgtcinsertpkg

\title{Metric Design != Metric Behavior: Improving Metric Selection \\for the Unbiased Evaluation of Dimensionality Reduction}

\author{Jiyeon Bae\thanks{e-mail: bjy7266@gmail.com}\\ %
        \scriptsize Seoul National University %
\and Hyeon Jeon\thanks{e-mail: hj@hcil.snu.ac.kr}\\ %
     \scriptsize Seoul National University %
\and Jinwook Seo\thanks{e-mail: jseo@snu.ac.kr, corresponding author}\\ %
      \scriptsize Seoul National University}

\abstract{
Evaluating the accuracy of dimensionality reduction (DR) projections in preserving the structure of high-dimensional data is crucial for reliable visual analytics. 
Diverse evaluation metrics targeting different structural characteristics have thus been developed.
However, evaluations of DR projections can become biased if highly correlated metrics---those measuring similar structural characteristics---are inadvertently selected, favoring DR techniques that emphasize those characteristics.
To address this issue, we propose a novel workflow \revise{that reduces bias in the selection of evaluation metrics} by clustering metrics based on their empirical correlations rather than on their intended design characteristics alone. 
Our workflow works by computing metric similarity using pairwise correlations, clustering metrics to minimize overlap, and selecting a representative metric from each cluster.
Quantitative experiments demonstrate that our approach \revise{improves the stability of DR evaluation, which indicates that our workflow contributes to mitigating evaluation bias.}

} % end of abstract

\keywords{Dimensionality reduction, Evaluation metrics, Correlation analysis, Benchmarking, Visual analytics}

\begin{document}


\firstsection{Introduction}

\maketitle

Dimensionality Reduction (DR) techniques play a central role in the visual analytics of high-dimensional data across diverse domains, including bioinformatics \cite{ivan13}, HCI \cite{Cavallo18}, and signal processing \cite{Rui16}.
These techniques aim to project high-dimensional data into lower dimensions while preserving key structural characteristics, such as cluster formations or neighborhood relationships.
% However, due to the inherent limitations of projecting complex structures from high-dimensional spaces to fewer dimensions, DR projections inevitably emphasize certain structural properties at the expense of others---each having its own focused characteristics. 
However, inherent limitations in projecting complex high-dimensional structures to lower dimensions cause DR projections to inevitably emphasize certain structural properties at the expense of others---each having its own focused characteristics. Therefore, quantitatively evaluating DR projections to understand those focused characteristics is essential for interpreting their reliability and limitations before applying them in visual analytics tasks \cite{espadoto21tvcg, jeon2025arxiv, jeon25chi}.

To quantify the accuracy of DR projections comprehensively, researchers have developed various evaluation metrics, each designed to assess distinct structural characteristics of data projections, e.g., local neighborhood preservation \cite{venna06} or global distances between points \cite{kruskal64, kruskal64psy}. Although diverse metrics help capture different structural characteristics, the selection of evaluation metrics itself significantly influences evaluation outcomes. In particular, choosing highly correlated metrics---metrics assessing very similar structural properties---can bias evaluations toward DR techniques that specifically optimize for these characteristics\revise{, i.e., can erroneously favor certain DR techniques as consistently superior to others.} For instance, predominantly using metrics sensitive to local neighborhood structures may disproportionately favor methods like t-SNE or UMAP, which optimize projections focusing on local relationships \revise{(see Appendix A)}. 
Currently, a common practice to avoid evaluation bias is selecting metrics based on their stated design goals (e.g., optimizing local, cluster-level, or global). 
However, such an approach can be biased because metrics designed with different intentions may still behave similarly in practice and vice versa. Thus, there still remains a need for an empirically driven approach \revise{that reduces bias in the selection of DR evaluation metrics.}

To address this problem, we introduce a workflow \revise{that selects a subset of evaluation metrics according to their empirical behavior.}
Our workflow consists of three main steps: (1) computing empirical correlations among metrics across diverse DR projections; (2) clustering these metrics using their correlation-based similarity; and (3) selecting representative metrics from each cluster to minimize redundancy and bias.
%\revise{Quantitative experiments show that our approach improves the stability of DR evaluation---that is, the consistency of the rankings of DR techniques across different metric sets---and outperforms baseline methods. These results verify that our workflow contributes to reducing evaluation bias.}
\revise{Quantitative experiments show that our approach improves stability of DR evaluation, i.e., the consistency of the rankings of DR techniques across different metric sets, outperforming baseline methods. The results thus verify that our workflow contributes to reducing evaluation bias.}

\section{Background and Related Work}

We review prior studies on the evaluation of DR techniques and discuss existing approaches to evaluation metric selection.

% recent work that comprehensively unifies local, cluster-level, and global distortion metrics for evaluating dimensionality reduction. As distortions in low-dimensional embeddings can seriously undermine the reliability of subsequent data analysis, a wide range of measures has been devised to rigorously quantify embedding fidelity.

% DR 기법 사용하는 예시 논문 더 추가

\subsection{Evaluating Dimensionality Reduction Techniques}

\label{sec:evaltech}
% - Classical Methods (PCA, MDS, etc.), Nonlinear Approaches (Isomap, LLE, t-SNE, UMAP)
% - Traditional Metrics (Stress, Trustworthiness, Continuity), Local vs. Global vs. Cluster Structural Emphasis, Common Use Cases Across Application Domains

% zadu \cite{jeon23vis}

% A critical step in using dimensionality reduction (DR) techniques is to \emph{evaluate how faithfully the resulting low-dimensional embeddings preserve the structure} of the original high-dimensional data. 
% Researchers have developed diverse evaluation metrics to measure the accuracy of DR projections.
% These metrics are largely divided into three classes: \textit{local}, \textit{cluster-level}, and \textit{global} metrics \cite{jeon23vis}.
Researchers have developed diverse evaluation metrics to measure the accuracy of DR projections.
These metrics are broadly categorized into three classes: \textit{local}, \textit{cluster-level}, and \textit{global} metrics \cite{jeon23vis}.
% from multiple perspectives, ranging from local neighborhood preservation to cluster-level consistency and global distance fidelity.\cite{jeon23vis}

\paragraph{Local metrics}
These metrics focus on evaluating how well DR projections preserve local neighborhood structure.
For example, \textit{Trustworthiness \& Continuity (T\&C)} \cite{venna06} and \textit{Mean Relative Rank Error (MRRE)} \cite{john09} penalize a projection in which the nearest neighbors in the original space are no longer neighbors in the projection, or vice versa.

% Several DR evaluation methods focus on \emph{local neighborhood preservation}, examining how accurately the $k$-nearest neighbors in the original data remain neighbors in the low-dimensional embedding. For instance, \textit{Trustworthiness \& Continuity (T\&C)}\cite{venna06} and \textit{Mean Relative Rank Error (MRRE)}\cite{john09} penalize points whose neighbors shift or disappear in the projection, while \textit{Local Continuity Meta-Criteria (LCMC)}\cite{chen09} focuses on the proportion of true neighbors that persist across both spaces. Other approaches include \textit{Neighborhood Hit}\cite{paulovich08}, which checks how often neighbors share the same class label, and \textit{Procrustes Measure}\cite{goldberg08}, which assesses local geometric alignment. Meanwhile, \textit{Neighborhood Dissimilarity}\cite{fujiwara23pavis} compares shared-nearest-neighbor graphs in high-dimensional and projected spaces, spotlighting subtle structural distortions, whereas \textit{Class-Aware T\&C}\cite{fujiwara23pavis} extends T\&C by distinguishing inter-class from intra-class neighbor mismatches.

\paragraph{Cluster-level metrics}
These metrics evaluate whether projections accurately preserve the cluster structure of the original data. 
For example, \textit{Distance Consistency} \cite{sips09} or clustering validation measures like \textit{Silhouette} \cite{joia11} measure how well the labeled classes stay separated in projections, based on the assumption that these classes are well separated in the high-dimensional space.

% Beyond local neighborhoods, \emph{cluster-level metrics} probe whether more aggregated or class-based structures remain intact. For instance, \textit{Steadiness \& Cohesiveness (S\&C)}\cite{Jeon22tvcg} evaluates whether points that cluster together in the original space remain close in the low-dimensional embedding (and vice versa), highlighting missing or distorted groups. \textit{Distance Consistency}\cite{sips09} checks how reliably labeled classes stay separated following projection, while \textit{Internal Clustering Validation Measures}\cite{joia11} assess how coherent partitions are within the embedded space itself. In addition, combining clustering outputs with \textit{External Clustering Validation Measures}\cite{xiang21} quantifies how well known labels or categories are preserved in the lower-dimensional representation.

\paragraph{Global metrics}
Finally, global metrics examine whether global relationships like pairwise distances between data points or clusters of the original data remain consistent in the low-dimensional projection.
For instance, \textit{Stress} \cite{kruskal64, kruskal64psy} measures discrepancies between the distance matrices of the original and projected spaces, whereas \textit{Kullback--Leibler (KL) Divergence} \cite{hinton02} evaluates differences in how the probability distributions vary across these two spaces.

% closely the rankalign with those in the original space.
% At the global level, \emph{global metrics} examine whether large-scale relationships from the original high-dimensional data remain consistent in the low-dimensional embedding. 
% For instance, \textit{Stress}\cite{kruskal64, kruskal64psy} measures discrepancies between the original and embedded distance matrices, whereas \textit{Kullback--Leibler Divergence (KL Divergence)}\cite{hinton02} evaluates differences in estimated probability distributions across the two spaces. \textit{Distance-to-Measure}\cite{chazal11} sums deviations in local densities, and \textit{Topographic Product}\cite{bauer92} checks how faithfully topological relationships are preserved following projection. In addition, correlation-based measures such as \textit{Pearson’s $r$}\cite{geng05} and \textit{Spearman’s $\rho$}\cite{whitfield18} quantify how closely the embedded distances (or their ranks) align with those in the original space.

% \subsection{Evaluation Metrics for DR}

\vspace{4pt}
\noindent 
\textit{Our contribution.}
Although various DR evaluation metrics have been developed, the community lacks systematic approaches for selecting the optimal set of metrics. Evaluations of DR projections might thus inadvertently emphasize certain structural characteristics disproportionately, potentially misguiding users with biased projections. \revise{We address this problem by introducing a workflow that selects evaluation metrics that have dissimilar behavior.}

%(rebutt) We address this problem by introducing a workflow that selects a mutually dissimilar---and therefore unbiased---set of metrics.

% Although various DR evaluation metrics have been developed, the community lacks systematic approaches for selecting the optimal set of metrics that minimizes redundancy and bias. 
% Evaluations of DR projections might thus inadvertently emphasize certain structural characteristics disproportionately, potentially misguiding users with biased projections.
% We address this problem by introducing a workflow that selects a mutually dissimilar set of metrics.

% Current DR evaluation practices frequently include sets of highly correlated metrics, which can lead to redundant assessments that disproportionately reward certain structural aspects (e.g., local neighborhood preservation) while neglecting others (e.g., global distance fidelity). Such metric-set bias skews performance comparisons and obscures the full potential of competing DR methods. To address this, we introduce a workflow that quantifies pairwise correlations between DR evaluation metrics, then clusters them by similarity and selects centroid metrics from each cluster. This strategy curtails the dominance of overrepresented features and fosters a more balanced assessment of DR performance.

\subsection{Existing Approaches for Metric Selection}

% Although the community lacks a standardized way to select DR evaluation metrics, researchers commonly select metrics considering their intended target characteristics (\autoref{sec:evaltech}).
% For example, Espadoto et al. \cite{espadoto21tvcg} leverage both local and global metrics to evaluate the accuracy of DR techniques comprehensively. 
% Similarly, several studies proposing new DR techniques \cite{vikram19icoei, jeon22vis, moor20iclm} employ both local and global metrics. 
% Van der Maaten et al. \cite{vandermaaten09} exclusively employ local metrics, which they categorize by the degree of locality each targets; they argue that the metrics complement each other in capturing aspects of manifold preservation.
Although the community lacks a standardized way to select DR evaluation metrics, researchers commonly select metrics considering their intended target characteristics (\autoref{sec:evaltech}).
For example, Espadoto et al. \cite{espadoto21tvcg} leverage both local and global metrics to evaluate the accuracy of DR techniques comprehensively. 
% Similarly, several studies proposing new DR techniques or benchmarking DR techniques \cite{jeon22vis, moor20iclm, amid2022, wang2021, atzberger24} employ both local and global metrics to evaluate the techniques.
% Similarly, several studies proposing new DR techniques or benchmarking DR techniques \cite{jeon22vis, moor20iclm, amid2022, wang2021, atzberger24} employ both local and global metrics to evaluate the techniques.
Similarly, prior research proposing new DR techniques \cite{jeon22vis, moor20iclm, amid2022, wang2021} employ both local and global metrics, \revise{while benchmark studies compare DR techniques across diverse structural levels \cite{atzberger24, atzberger23, atzberger24i}.}
% Van der Maaten et al. \cite{vandermaaten09} exclusively employ local metrics, which they categorize by the degree of locality each targets; they argue that the metrics complement each other in capturing aspects of manifold preservation.
%(Rebutt) Van der Maaten et al. \cite{vandermaaten09} employ only local metrics, each targeting a distinct aspect of neighborhood preservation, and argue that these metrics provide a comprehensive view of manifold preservation.

\revise{However, several studies show that such selection approaches cannot ensure a fair assessment of projection quality.}
 \revise{Thrun et al. \cite{Thrun23} identify bias in unsupervised DR evaluation metrics by using graph theory, demonstrating that each metric fails to evaluate structure preservation correctly when the input data violates its own structural assumptions.} 
 \revise{Machado et al. \cite{machado25} find that adversarially optimized projections can inflate correlated quality scores and construct a guardrail set of DR metrics through correlation analysis and clustering to address this issue.}

\vspace{4pt}
\noindent
\textit{Our contribution.}
\revise{Our proposed workflow empirically clusters metrics according to their actual observed behaviors, ensuring the selected metrics offer complementary rather than redundant evaluations. This is done by evaluating the behavior of DR metrics across the projections generated by 96 datasets and 40 DR techniques. The recommended compact yet diverse set of metrics for DR evaluation also reduces unnecessary computational costs. Our approach systematically addresses the critical shortcomings of existing DR evaluation practices while complementing them.}

\section{The Workflow}
% \section{The Workflow for Finding Optimal Metric Clusters}

% We introduce our workflow to find an unbiased set of metrics. 
% We then discuss the technical details on how we implement the workflow. 

\label{sec:workflow}

% \subsection{Workflow Design}

We propose a workflow for selecting DR evaluation metrics that minimizes redundancy and bias by focusing on their empirical behavior.
Our workflow begins by computing pairwise correlations between metrics as a proxy for similarity. 
We then cluster the metrics according to this similarity to identify optimal groups.
Finally, we select representative metrics with minimal pairwise similarity. 
Please refer to Appendix B for the evaluation metrics \revise{and parameters} we use.

% \subsection{Computing Correlations between Metrics}
\paragraph{(Step 1) Computing pairwise correlations}
% We construct a similarity matrix where each row and column represent individual evaluation metrics, and the cell value represents the similarity between the metrics corresponding to the row and column. 
% We aim to reflect the degree to which two metrics behave similarly in practice through their similarity.
% To do so, we first prepare 96 high-dimensional datasets with varying numbers of points, dimensions, and distributions \cite{jeon25tpami}.
% Then, for each dataset, we create 300 diverse projections with varying visual patterns. This is done by repeating the procedure of randomly selecting one of the 40 DR techniques and sampling its hyperparameters from predefined random ranges (see the list of techniques and hyperparameter ranges in Appendix A). 
% Afterwards, again for each dataset, we evaluate the accuracy of the corresponding projections using evaluation metrics.
% Finally, we compute the correlation between rankings made by evaluation metrics using Spearman’s $\rho$ for each dataset and determine the similarity of two metrics as the average of their correlations across the entire set of datasets. 
% Spearman correlation is used because it robustly captures monotonic relationships, even when metrics are non-linearly related.
We construct a similarity matrix in which each row and column corresponds to an evaluation metric, and each cell stores the similarity between each pair of metrics.
We aim to reflect how closely two metrics behave through their similarity. 
To obtain it, we first prepare 96 high-dimensional datasets that vary in size, dimensionality, and distribution \cite{jeon25tpami}.
For each dataset, we then produce 300 diverse projections exhibiting different visual patterns. 
% We achieve this by repeatedly selecting one of 40 DR techniques at random and randomly sampling its hyperparameters from predefined ranges (see the list of techniques and hyperparameter ranges in Appendix A). 
We achieve this by repeatedly selecting one of the 40 DR techniques at random and sampling its hyperparameters from predefined ranges (see Appendix C). 
For every dataset, we quantify the quality of each projection using the chosen evaluation metrics. 
We subsequently rank the 300 projections for each metric, compute \revise{Spearman’s rank correlation coefficient ($\rho$) between all pairs of rankings within each dataset}, and define the similarity between two metrics as the average of these correlations across all datasets. We employ \revise{Spearman’s $\rho$ because it compares the ranks of metric values, enabling it to capture monotonic relationships even under nonlinear distribution of metric scores \cite{Spearman1987}.}

% We subsequently rank the 300 projections for each metric, compute Spearman’s $\rho$ between all pairs of rankings per dataset, and define the similarity between two metrics as the average of these correlations across all datasets.

% We employ Spearman correlation because it reliably captures monotonic relationships, even when metrics are non-linearly related.

\paragraph{(Step 2) Clustering metrics}
% We cluster the evaluation metrics to prevent the selection of measures that accentuate identical structural characteristics.
% We convert the similarity matrix to a distance matrix by subtracting each entry from one.
% We then apply hierarchical clustering to the metrics using this distance matrix as input \cite{Murtagh11,Nielsen2016}.
% Hierarchical clustering is chosen for its robustness to noise and its stability across repeated runs \cite{Jain99}.
We cluster the evaluation metrics to avoid selecting measures that emphasize identical structural characteristics. 
We convert the similarity matrix to a distance matrix by subtracting each entry from 1.
We then apply hierarchical clustering \cite{Murtagh11,Nielsen2016} to the metrics, using this distance matrix as input \revise{and employing average linkage \cite{Moseley23,laber2024} for merging clusters.}
Hierarchical clustering is chosen for its robustness to noise and stability across multiple runs \cite{Jain99}.

\paragraph{(Step 3) Selecting representative metrics}
% From each cluster, we compute the average similarity between each metric and all other metrics within the cluster. We then select the metric with the highest average similarity to other cluster members as the representative. 
For each cluster, we compute the average similarity of every metric to all other metrics in the cluster and select the metric with the highest average similarity as the representative. This strategy ensures that the selected metrics best represent the characteristics of each cluster while minimizing overlap across the full set.

% For each cluster, we compute the average similarity of each metric to all other metrics within the cluster. We then select the metric with the highest average similarity to the other cluster members as the representative. This strategy ensures that selected metrics best represent the characteristics of their respective clusters while maintaining minimal overlap across the full set.

\vspace{4pt}
% In summary, our workflow systematically selects an unbiased and diverse set of evaluation metrics by empirically analyzing their behavior, clustering based on practical similarity, and carefully selecting representatives to minimize redundancy.
% (Rebutt) In summary, our workflow selects an unbiased and diverse set of evaluation metrics by analyzing their behavior, clustering based on similarity, and selecting representatives to minimize redundancy.
In summary, our workflow \revise{aims to alleviate bias in selecting a diverse set of evaluation metrics by analyzing their behavior and clustering them based on similarity to minimize redundancy.}

\begin{figure}[t]
    \captionsetup{belowskip=-11pt}
    \centering
    \includegraphics[width=\linewidth]{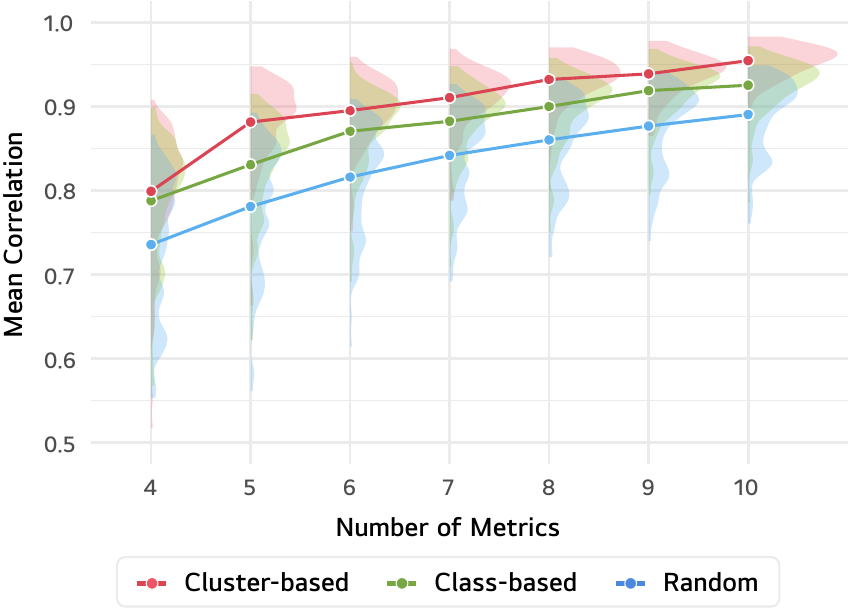}
    \caption{%
        Effect of metric selection strategy on rank stability of DR techniques.  
        % For each number of selected metrics $k$ ($4\le k\le10$), half‑violin “slab” plots depict the distribution of Spearman rank correlations obtained from 200 independent metric sets evaluated on 96 datasets.  
        % Filled circles denote the grand mean for each strategy and are connected by solid lines (blue = Random, green = Class‑based, red = Cluster‑based).  
        The cluster‑based approach yields consistently higher stability for $k \ge 5$, indicating that this strategy mitigates evaluation bias more effectively than random or class‑balanced sampling.  
        % An ANCOVA with $k$ as covariate confirms a significant main effect of strategy on stability ($F_{2,2012}=315.8,\;p<.001$), supporting the visual trend.%
    }
    \label{fig:rank}
\end{figure}

\begin{figure*}[t!]
    \captionsetup{belowskip=-11pt}
    \centering
    \includegraphics[width=1\linewidth]{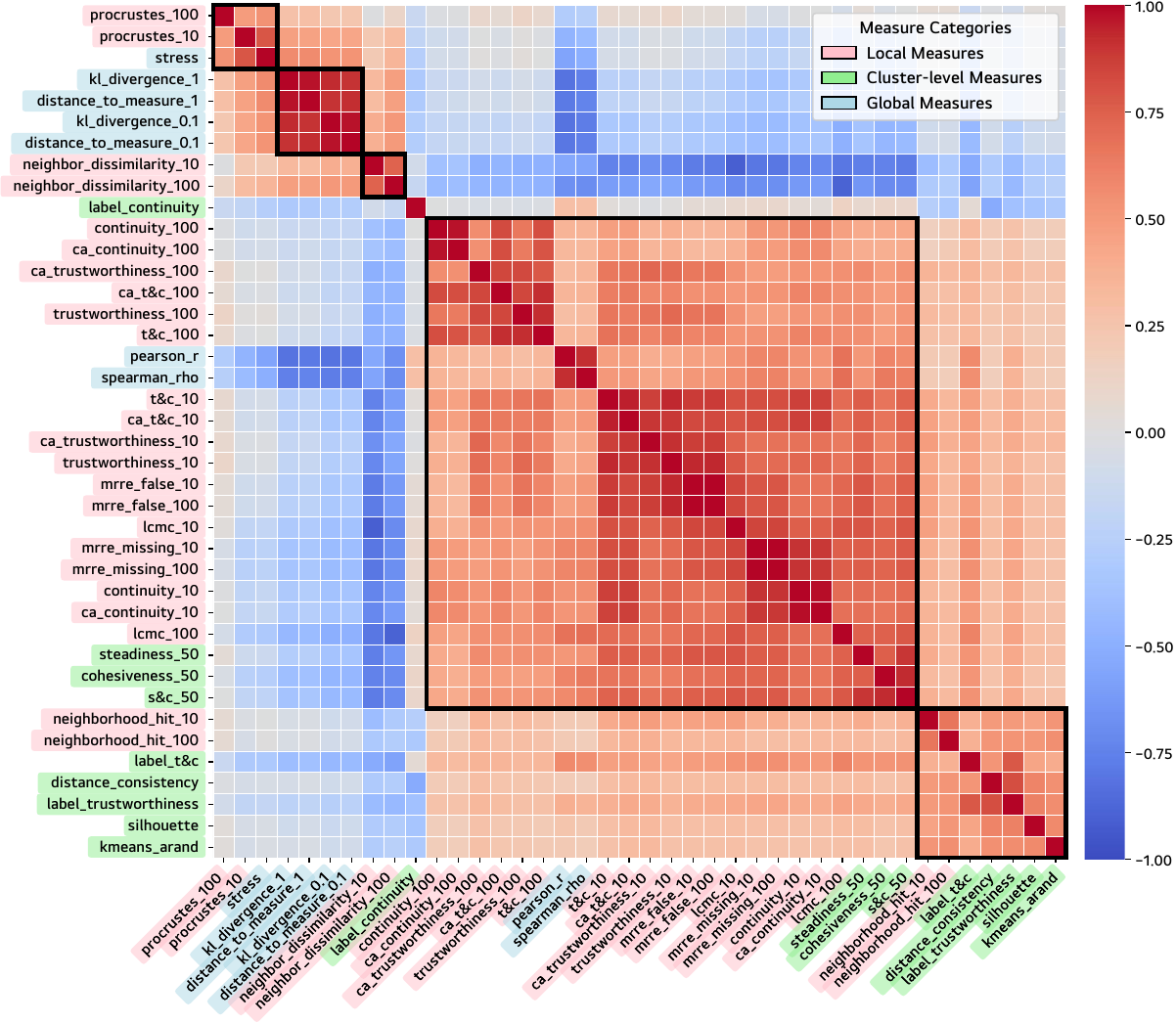}
    \vspace{-11pt}
    \caption{%
    % A heatmap that visualizes the matrix that visualizes the Spearman correlations between DR evaluation metrics.
    Heatmap of the Spearman correlation matrix among DR evaluation metrics. The order of metrics is determined by hierarchical clustering. Black borders denote the five optimal clusters obtained by our workflow. \revise{Singleton clusters are removed as outliers that could add noise to DR evaluation (see Appendix D).}
    Row and column labels are color‑coded according to their original design categories---local (pink), cluster‑level (light green), and global (light blue)---highlighting that metrics with disparate design intentions can exhibit highly similar empirical behavior and thus group together. The numeric suffixes on local metrics indicate the number of $k$-nearest neighbors considered, whereas those on global metrics denote the bandwidth used to compute similarity between data points, commonly written as $\sigma$ \cite{jeon23vis}.
        % The block‑diagonal structure confirms that the selected representative metric from each cluster will minimise redundancy in downstream evaluations.%
    }
    \label{fig:heatmap}
\end{figure*}

\section{Evaluation}

\label{sec:evaluation}
% We evaluate whether our workflow can reduce the bias in DR evalaution. To doso, stability ...

% We aim to investigate whether our workflow can effectively mitigate bias in DR evaluation. In particular, we assess the stability and consistency of the ranking outcomes under different metric selections and clustering approaches.

% \subsection{Objectives and Design}

% Metric Sampling Approaches, Pairwise Correlation Analysis --> 위에 instaitnaition 설명을 따른다.

% \subsubsection{Metric Sampling Approaches}
% \subsubsection{Pairwise Correlation Analysis}

% \subsection{Results and Discussions}

% \subsection{}
% \subsection{}
% \subsection{Ranking DR Methods}
% \subsection{Results}
% - include visualization

We evaluate the effectiveness of our workflow in selecting a set of evaluation metrics that minimizes bias in DR evaluation. 

\subsection{Objectives and Design}
% \subsection{Evaluation Objectives}

Our goal is to verify whether our workflow mitigates bias compared to baseline strategies for metric selection. 
We compare three strategies for selecting DR evaluation metrics: \textbf{Random} selection, \textbf{Class-based} selection, and \textbf{Cluster-based} selection.
Random selection refers to a strategy in which evaluation metrics are randomly drawn from the available metrics. 
Class-based selection distributes an equal share of metrics across each category (local, cluster-level, and global) on average.
Cluster-based selection randomly picks a single metric from each cluster produced by our workflow.

% Random selection refers to a strategy in which evaluation metrics are randomly drawn from the available metrics. 
% Class-based selection distributes an equal share of metrics to each category (local, cluster-level, and global) on average.
% Cluster-based selection randomly picks a single metric from each cluster produced by our workflow.

% In contrast, class-based selection distributes an equal share of metrics to each category (local, cluster-level, and global) on average.
% Finally, cluster-based selection randomly picks a single metric from each cluster produced by our workflow. 
% In contrast, class-based selection picks an equally distributed number of metrics from each category of evaluation metrics (local, cluster-level, and global). 
\paragraph{Procedure}
% We evaluate the degree to which each selection strategy is biased by the variability of evaluation outcomes across multiple executions.
We assess each selection strategy’s bias by measuring the variability of evaluation outcomes across multiple executions.
% The rationale is that if a strategy is biased, the structural characteristics emphasized by the chosen metrics vary widely across runs, causing the overall ranking of projections to fluctuate.
The rationale is that if a strategy is biased, the structural characteristics emphasized by the chosen metrics vary widely across runs, causing the overall ranking of projections to fluctuate.

% The rationale underlying this design is that if a selection strategy is biased, the structural characteristics emphasized by the chosen metrics vary widely across runs, and the overall ranking of projections fluctuates accordingly.
% The rationale is that, if a strategy is biased, the distribution of structural characteristics captured by the selected metrics will vary widely between runs, inducing corresponding fluctuations in the evaluation results.
% The rationale based on this desing is that If a selection strategy is biased, the distribution of structural characteristics emphasized by the chosen metrics varies widely across runs, and the overall ranking of projections fluctuates accordingly.
We first prepare 96 datasets and generate 300 projections by randomly sampling DR techniques and hyperparameter settings. This process is identical to the procedure for generating projections in Step 1 of our workflow (\autoref{sec:workflow}). Second, for each selection strategy, we execute it 200 times, yielding 200 distinct sets of evaluation metrics. Finally, for each dataset, we quantify how much the rankings of the 300 projections vary across metric sets by computing the pairwise Spearman correlation $\rho$ between the rankings produced by each metric set.
The final correlation for each dataset is obtained by averaging all pairwise correlations. \revise{We then average these dataset-level correlations across all datasets to obtain the rank stability for each selection strategy.} We repeat this process while increasing the number of clusters from 4 to 10.

\subsection{Results and Discussions}
\label{sec:results}
% Our results (\autoref{fig:enter-label}) find that the cluster-based selection strategy shows the highest rank stability between different trials, outperforming baselines (class-based and random selections).
% We observe a trend where the cluster-based selection strategy consistently yields higher rank stability across different cluster numbers $k$. 
% To confirm this observation, we run a one–way ANCOVA, treating \(k\) as a covariate and
% \textit{selection strategy} as a fixed factor.
% The results confirm that three different conditions have significantly different rank stability (\(F_{2,2012}=315.84,\; p<.001\)).

Our results (\autoref{fig:rank}) show that the cluster‑based selection strategy achieves the highest rank stability across trials, outperforming the class‑based and random baselines. 
We observe that this advantage persists across all examined cluster counts \(k\). 
To test this observation, we conduct a one‑way ANCOVA, treating \(k\) as a covariate and \textit{selection strategy} as a fixed factor. 
The analysis reveals significant differences in rank stability among the three strategies (\(F_{2,2012} = 315.84,\; p < 0.001\)). 

To further investigate this finding, we conduct a one‑way ANOVA to compare rank stability among the three selection strategies for each \(k\).
The analysis reveals significant effects of selection strategy for \(5 \le k \le 10\) (\(p < 0.001\) for all).
Bonferroni‑corrected post‑hoc comparisons indicate that the cluster‑based selection strategy achieves higher rank stability than the other two strategies in these cases (\(p < 0.001\) for all).
However, no significant differences are observed at \(k = 4\) \((F_{2,285} = 1.26,\; p = 0.285)\).
% To further investigate this phenomenon, we conduct a one‑way ANOVA to compare rank‑stability differences among the three conditions for each \(k\).
% The analysis reveals significant between‑condition effects for \(5 \leq k \leq 10\) (\(p<.001\) for all).
% Bonferroni‑corrected post‑hoc comparisons indicate that the cluster‑based selection strategy attains higher rank stability than the other two strategies in these cases (\(p<.001\) for all).
% However, no significant difference is observed at \(k = 4\) (\(F_{2,285}=1.26,\; p=.285\)).

\revise{These results show that our workflow mitigates bias in DR evaluation more effectively than the baselines, given the same number of metrics.} This enables practitioners to obtain reliable assessments, reducing computational overhead.

% (rebutt) These results show that our workflow mitigates bias in DR evaluation, even with a small set of metrics. This enables practitioners to obtain reliable assessments, reducing computational overhead.

% These results confirm that our workflow effectively mitigates bias in DR evaluation, even with a minimal metric set. Therefore, practitioners can achieve unbiased assessments with reduced computational overhead.

% revealed a
% significant main effect of the strategy on rank stability
% (\(F=358.4,\; p=5.11\times10^{-120}\)).

% To further examine this phenomenon, we also run a one-way ANOVA to compare the differences in rank stability between the three conditions for each $k$. As a result, we find a significant difference between conditions for $5 \leq k \leq 10$ ($p < .001$ for all). Post-hoc analysis using Bonferroni correction reveals that the cluster-based selection strategy has significantly high rank stability compared to the other two conditions for these cases ($p < .001$ for al ).
% However, we find no no significance for $k=4$ (\(F_{2,285}=19.06,\; p<.001\)).

\begin{figure}[t]
    \captionsetup{belowskip=-11pt}
    \centering
    \includegraphics[width=\linewidth]{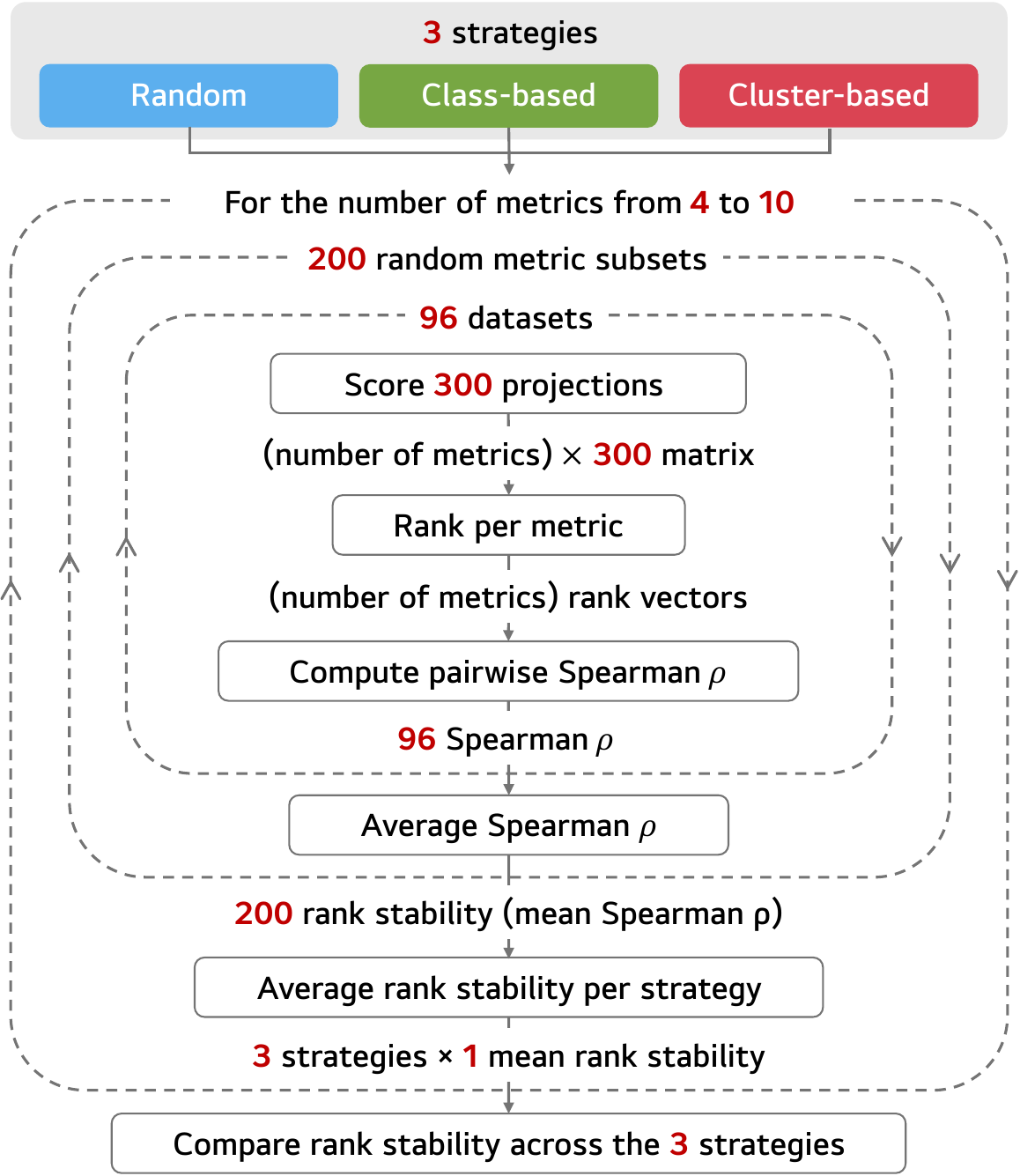}
    \caption{%
        Procedure for evaluating rank stability across metric selection strategies. The results are depicted in \autoref{fig:rank}.
        % The results of this evaluation are depicted in \autoref{fig:rank}.
        % For each value of the number of metrics \(m\in\{4,\dots,10\}\) three strategies—Random, Class‑based, and Cluster‑based—each sample \(m\) metrics 200 times. 
        % Every metric set scores the same 300 projections on 96 datasets, converts the scores into \(m\) rank vectors, and averages all pairwise Spearman correlations to obtain one stability score. 
        % The resulting 200 stability scores for every \(m\)–strategy pair are later aggregated for statistical comparison.%
    }%
    \label{fig:workflow}
\end{figure}

\section{Recommendations}
% In previous sections, we introduced a workflow (\autoref{sec:workflow}) to obtain unbiased sets of DR evaluation metrics and empirically demonstrated its effectiveness (\autoref{sec:evaluation}).
% Here, we apply this workflow to recommend a compact and unbiased set of evaluation metrics.
% We first describe the selection procedure and then discuss the resulting recommendations.
In previous sections, we introduce \revise{a workflow that mitigates bias in the selection of DR evaluation metrics} (\autoref{sec:workflow}) and validate it empirically (\autoref{sec:evaluation}).
Here, we apply this workflow to recommend a compact \revise{set of evaluation metrics.}
We first describe the selection procedure and then discuss the resulting recommendations.
% We first describe the selection procedure and then discuss the resulting recommendations.
% Here, we apply this workflow to recommend a compact and unbiased set of evaluation metrics.
% We first describe the selection procedure and then discuss the resulting recommendations.
% In the previous sections, we have established a workflow (\autoref{sec:workflow}) to find aㄱn unbiased set of DR evaluation metrics and empirically demonstrate its effectiveness (\autoref{sec:evaluation}).
% Here, we apply this workflow to recommend a set of evaluation metrics that are unbiased and can be efficiently computed.
% We first describe the selection procedure and then discuss the resulting recommendations.

% Objective: workflow가 있고, 우리가 실험을 해서 이게 좋다는 것을 밝혔다.
% 그래서 우리는 여기서 우리의 workflow를 활용해서 최종적으로 unbiased 하면서 efficeint하게 계산할 수 있는 set of metric을 추천하고자 한다. 우리는 먼저 그 procedure를 설명하고 결과를 discuss한다. 

\subsection{Procedure}
% Following our workflow, we generate 300 projections for each of the 96 benchmark datasets by randomly selecting one of 40 DR techniques and sampling its hyperparameters.
% We rank these projections using every evaluation metric and construct a similarity matrix, whose entries represent the mean Spearman correlations between each pair of metrics across all datasets.
We execute our workflow using the evaluation metrics employed in our evaluation (\autoref{sec:evaluation}).
To determine the optimal metric count, we measure the diversity of the selected metrics as the cluster count increases.
We then identify the optimal cluster count using the elbow method.
Details of both procedures appear below.

% We then apply average--linkage hierarchical clustering to this matrix.
% Within each cluster, we designate as representative the metric with the highest mean similarity to the other members. To assess how well the representatives cover distinct structural perspectives and to determine how many clusters to retain, we quantify their \emph{diversity} and identify the optimal cluster count via an elbow analysis, as summarized below.

\paragraph{Computing diversity}
We define diversity as the minimum dissimilarity between each metric and its nearest neighbor in the selected set.
First, for a set of $k$ metrics $\{M_1, M_2, \ldots, M_k\}$, we quantify the degree to which each metric is distinct from the others:
\setlength{\abovedisplayskip}{7pt}  % 수식 위 여백
\setlength{\belowdisplayskip}{4pt}  % 수식 아래 여백
\[
\operatorname{Ind}(M_i)=\min_{j \neq i}\bigl(1 - R_{i,j}\bigr),
\]
where $R_{i,j}$ denotes the Spearman correlation between $M_i$ and $M_j$, so $\operatorname{Ind}(M_i)$ measures how far $M_i$ lies from its nearest neighbor among the other representative metrics. 

We then compute the diversity of the set as:
\setlength{\abovedisplayskip}{4pt}  % 수식 위 여백
\setlength{\belowdisplayskip}{4pt}  % 수식 아래 여백
\[
D=\sum_{i=1}^{k}\operatorname{Ind}(M_i).
\]
As $D$ typically grows with larger $k$, we use the normalized measure: $D_{\mathrm{norm}}=D/k$.

% We define diversity as the minimum dissimilarity between each metric and its closest neighbor in the selected set, ensuring that the chosen metrics capture distinct structural characteristics. Formally, for a selected set of $k$ metrics $\{M_1, M_2, \ldots, M_k\}$, we measure the degree to which each metric is distinct from the others:
% \[
% D_{\mathrm{norm}}=\frac{D}{k}.
% \]

\begin{figure}[t!]
    \captionsetup{belowskip=-11pt}
    \centering
    \includegraphics[width=\linewidth]{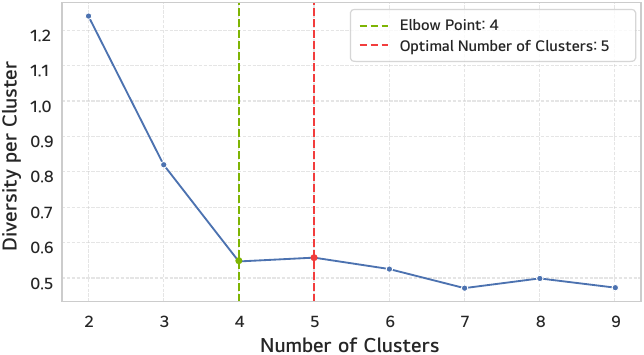}
    \caption{%
        Elbow analysis for selecting the number $k$ of metric clusters.  
        \revise{We adopt $k=5$, the largest cluster count beyond the elbow (red dashed line), as the optimal number of clusters (see Appendix E).}
    }
    \label{fig:recommendation}
\end{figure}

\paragraph{Determining the optimal number of clusters}
We observe that diversity tends to rise with the cluster count $k$ but eventually levels off.
We therefore apply the \emph{elbow method} to identify the cluster count beyond which additional clusters yield negligible diversity gains.
We employ Kneedle \cite{satopaa11icdcws}, which automatically detects the elbow.
We choose the $k$ just beyond the elbow that maximizes diversity among the representative metrics.
This choice balances diversity against fragmentation, ensuring each representative metric captures a distinct structural perspective.

% to balance diversity against fragmentation, thereby ensuring that each representative metric captures a distinct structural perspective.

% By plotting $D_{\mathrm{norm}}$ against the number of clusters~$k$, we observe how rapidly diversity grows as~$k$ increases.
% To determine the optimal~$k$, we employ the \emph{elbow method} \cite{Yong21,Kumar14}, a common clustering heuristic: we look for an ``elbow'' in the curve that marks the point beyond which additional clusters yield diminishing returns in $D_{\mathrm{norm}}$.
% We locate this elbow automatically by evaluating the curvature of the $D_{\mathrm{norm}}$ curve and selecting the first pronounced knee.
% We then choose the largest~$k$ beyond this elbow to balance diversity against fragmentation, thereby ensuring that each representative metric captures a distinct structural perspective.
% After clustering the metrics for a range of candidate values $k$, we designate exactly one \emph{representative item} per cluster by choosing the metric that exhibits the highest average correlation with other members in the same cluster. 

\subsection{Results and Recommendations}

% 이런 결과가 나왔다. 그래서 우리는 ~~게 뽑아서 평가하는 것을 추천하는데, label-continuity를 빼는걸 추천하는데 그 이유는 다 값

% We find that the optimal number of clusters is five (\autoref{fig:recommendation}).
% As shown in \autoref{fig:heatmap}, these clusters do not strictly align with the original classes of DR evaluation metrics—local, cluster-level, and global.
% For example, some cluster-level and global metrics are incorporated into the cluster mostly composed of local metrics (\autoref{fig:heatmap}, largest cluster).
% Such a result indicates that the classes of metrics do not precisely reflect their targeted structural characteristics, reaffirming the importance of our workflow in finding an unbiased set of metrics.
We find the optimal number of clusters is 5 (\autoref{fig:recommendation}).
% As shown in \autoref{fig:heatmap}, these clusters do not strictly align with the original classes of DR evaluation metrics—local, cluster-level, and global.
% For example, the largest cluster, predominantly composed of local metrics, also contains several cluster-level and global metrics (\autoref{fig:heatmap}).
As shown in \autoref{fig:heatmap}, the clusters do not strictly align with the original, design-based classes of DR evaluation metrics: local, cluster-level, and global.
For example, the largest cluster---although dominated by local metrics---also includes several cluster-level and global metrics (\autoref{fig:heatmap}).
This result indicates that original classes of metrics do not accurately reflect their intended structural characteristics, reaffirming the value of our workflow \revise{for reducing bias in metric selection.} The representative set comprises two local metrics (\texttt{neighbor\_dissimilarity} and \texttt{t\&c}), 
one cluster-level metric (\texttt{label\_trustworthiness}), 
and two global metrics (\texttt{stress} and \texttt{kl\_divergence}).
This set spans a broad range of structural characteristics with minimal redundancy, enabling practitioners \revise{to alleviate bias in DR evaluations.}

\section{Conclusion and Future Work}
% In this work, we address bias in DR evaluations caused by highly correlated metrics.
% We introduce a workflow that selects evaluation metrics with maximal mutual dissimilarity.
% Quantitative experiments demonstrate that our workflow substantially reduces evaluation bias relative to baselines, even with a small metric subset.
% Applying the workflow, we recommend a compact, diverse metric set---covering local, cluster-level, and global structures---to support more reliable and efficient DR evaluation in practice.

% While our work focuses on DR evaluation, metric redundancy and bias also appear in other machine-learning domains. For instance, widely used machine-translation evaluation metrics, such as BLEU \cite{mathur2020, Papin02}, ROUGE-L \cite{lin04, ganesan18}, and BERTScore \cite{zhang20}, are highly correlated and can bias system comparisons, and similar problems occur in evaluation of medical-image segmentation models \cite{muller2022}. 

% In the future, we plan to generalize our workflow for broader application across machine-learning and visualization evaluation tasks.
% Extending the approach to domains with heterogeneous, task-specific criteria poses challenges---for example, balancing semantic preservation and fluency in text generation, or managing trade-offs between segmentation accuracy and boundary precision in medical imaging.
% Ultimately, we envision empirically driven metric selection playing a critical role in ensuring fair and meaningful evaluation across diverse fields.

In this work, we propose a workflow that mitigates bias in DR evaluations caused by highly correlated metrics through the selection of metrics that are maximally dissimilar.
% In this work, we mitigate bias in DR evaluations caused by highly correlated metrics using a workflow that selects evaluation metrics with maximal mutual dissimilarity.
Quantitative experiments demonstrate that our workflow substantially reduces evaluation bias relative to baseline methods. Applying the workflow, we recommend a compact and diverse set of metrics to support more reliable and efficient DR evaluation in practice. 

\revise{However, our workflow is limited in that it cannot capture the full range of real-world data distributions or explore every possible hyperparameter configuration. In future work, we will examine whether our findings can be generalized by utilizing a larger set of datasets and DR techniques. }

% We also plan to examine whether the observed clusters stem from spurious correlations through qualitative analyses and, on the quantitative side, extensive hyper-parameter sweeps.

% We propose a workflow that mitigates bias in DR evaluations by selecting maximally dissimilar metrics. Experiments show that it reduces bias relative to baseline methods even with a small metric subset, leading us to recommend a compact, diverse set spanning local, cluster-level, and global structures.

% While our work focuses on DR evaluation, metric redundancy and bias also occur in other machine learning domains. For instance, widely used machine-translation evaluation metrics, such as BLEU \cite{mathur2020, Papin02}, ROUGE-L \cite{lin04, ganesan18}, and BERTScore \cite{zhang20}, are highly correlated and can bias system comparisons, and similar issues arise in medical image segmentation evaluation \cite{muller2022}. We plan to generalize our workflow for broader application across machine learning and visualization evaluation tasks.
% Extending the approach to domains with heterogeneous, task-specific criteria poses challenges---for example, balancing semantic preservation and fluency in text generation, or managing trade-offs between segmentation accuracy and boundary precision in medical imaging.
% Ultimately, we envision empirically driven metric selection playing a critical role in ensuring fair and meaningful evaluation across diverse fields.

\revise{We also aim to generalize our findings in other domains.}
Although we focus on DR evaluation, metric redundancy and bias also occur in other machine learning domains. For instance, \revise{in NLP domain, widely used machine translation metrics---such as BLEU \cite{Papin02, mathur2020}, ROUGE-L \cite{lin04, ganesan18}, and METEOR \cite{Lavie07}---are highly correlated and can bias system comparisons \cite{fabbri2021}.}  Similar issues arise in medical image segmentation evaluation \revise{\cite{muller2022, Taha2015}. The Dice similarity coefficient and Jaccard index are widely used but are mathematically related and were shown to produce identical rankings of segmentation methods \cite{Taha2015}.}
We plan to generalize our workflow to a broader range of machine learning and visualization evaluation tasks.
Extending the approach to heterogeneous, task-specific domains poses challenges---e.g., balancing semantic preservation with fluency in text generation, or reconciling segmentation accuracy with boundary precision in medical imaging.
Ultimately, we envision empirically driven metric selection playing a critical role in ensuring fair evaluation across diverse fields.

\acknowledgments{
This work was supported in part by the National Research Foundation of Korea (NRF),
funded by the Korean government (MSIT) under Grant 2023R1A2C200520911,
and in part by the Institute of Information and Communications Technology
Planning and Evaluation (IITP), funded by the Korean government (MSIT) under Grant RS-2021-II211343 [Artificial Intelligence Graduate School Program
(Seoul National University)]. The ICT at Seoul National University provided research facilities for this study.}
\bibliographystyle{abbrv-doi-narrow}

\bibliography{ref}

\begin{thebibliography}{10}
\renewcommand*{\sfdefault}{PTSansNarrow-TLF}

\bibitem{amid2022}
E.~Amid and M.~K. Warmuth.
\newblock Trimap: Large-scale dimensionality reduction using triplets, 2019. doi: \textsf{%
10\hspace{.1pt}\discretionary{.}{%
}{.}\hspace{.4pt}48550\discretionary{/}{%
}{/}ARXIV\hspace{.1pt}\discretionary{.}{%
}{.}\hspace{.4pt}1910\hspace{.1pt}\discretionary{.}{%
}{.}\hspace{.4pt}00204}


\bibitem{atzberger23}
Atzberger et~al.
\newblock Large-scale evaluation of topic models and dimensionality reduction methods for 2d text spatialization.
\newblock {\em IEEE Transactions on Visualization and Computer Graphics}, p. 1–11, 2023. doi: \textsf{%
10\hspace{.1pt}\discretionary{.}{%
}{.}\hspace{.4pt}1109\discretionary{/}{%
}{/}tvcg\hspace{.1pt}\discretionary{.}{%
}{.}\hspace{.4pt}2023\hspace{.1pt}\discretionary{.}{%
}{.}\hspace{.4pt}3326569}


\bibitem{atzberger24i}
D.~Atzberger et~al.
\newblock Quantifying topic model influence on text layouts based on dimensionality reductions.
\newblock In {\em Proceedings of the 19th International Joint Conference on Computer Vision, Imaging and Computer Graphics Theory and Applications}, vol. 1, GRAPP, HUCAPP and IVAPP, pp. 593--602, 2024. doi: \textsf{%
10\hspace{.1pt}\discretionary{.}{%
}{.}\hspace{.4pt}5220\discretionary{/}{%
}{/}0012391100003660}


\bibitem{atzberger24}
D.~Atzberger et~al.
\newblock A large-scale sensitivity analysis on latent embeddings and dimensionality reductions for text spatializations, Jan. 2025. doi: \textsf{%
10\hspace{.1pt}\discretionary{.}{%
}{.}\hspace{.4pt}1109\discretionary{/}{%
}{/}TVCG\hspace{.1pt}\discretionary{.}{%
}{.}\hspace{.4pt}2024\hspace{.1pt}\discretionary{.}{%
}{.}\hspace{.4pt}3456308}


\bibitem{Cavallo18}
M.~Cavallo and c.~Demiralp.
\newblock A visual interaction framework for dimensionality reduction based data exploration.
\newblock In {\em Proceedings of the 2018 CHI Conference on Human Factors in Computing Systems}, p. 1–13, 2018. doi: \textsf{%
10\hspace{.1pt}\discretionary{.}{%
}{.}\hspace{.4pt}1145\discretionary{/}{%
}{/}3173574\hspace{.1pt}\discretionary{.}{%
}{.}\hspace{.4pt}3174209}


\bibitem{espadoto21tvcg}
M.~Espadoto, R.~M. Martins, A.~Kerren, N.~S.~T. Hirata, and A.~C. Telea.
\newblock Toward a quantitative survey of dimension reduction techniques.
\newblock {\em IEEE Transactions on Visualization and Computer Graphics}, 27(3):2153--2173, 2021. doi: \textsf{%
10\hspace{.1pt}\discretionary{.}{%
}{.}\hspace{.4pt}1109\discretionary{/}{%
}{/}TVCG\hspace{.1pt}\discretionary{.}{%
}{.}\hspace{.4pt}2019\hspace{.1pt}\discretionary{.}{%
}{.}\hspace{.4pt}2944182}


\bibitem{fabbri2021}
Fabbri et~al.
\newblock Summeval: Re-evaluating summarization evaluation, 04 2021. doi: \textsf{%
10\hspace{.1pt}\discretionary{.}{%
}{.}\hspace{.4pt}1162\discretionary{/}{%
}{/}tacl\_a\_00373}


\bibitem{ganesan18}
K.~Ganesan.
\newblock Rouge 2.0: Updated and improved measures for evaluation of summarization tasks, 2018. doi: \textsf{%
10\hspace{.1pt}\discretionary{.}{%
}{.}\hspace{.4pt}48550\discretionary{/}{%
}{/}arXiv\hspace{.1pt}\discretionary{.}{%
}{.}\hspace{.4pt}1803\hspace{.1pt}\discretionary{.}{%
}{.}\hspace{.4pt}01937}


\bibitem{hinton02}
G.~Hinton and S.~Roweis.
\newblock Stochastic neighbor embedding.
\newblock In {\em Proceedings of the 16th International Conference on Neural Information Processing Systems}, NIPS'02, p. 857–864, 2002.

\bibitem{ivan13}
G.~Iván and V.~Grolmusz.
\newblock On dimension reduction of clustering results in structural bioinformatics.
\newblock {\em Biochimica et Biophysica Acta (BBA) - Proteins and Proteomics}, 1844(12):2277--2283, 2014. doi: \textsf{%
10\hspace{.1pt}\discretionary{.}{%
}{.}\hspace{.4pt}1016\discretionary{/}{%
}{/}j\hspace{.1pt}\discretionary{.}{%
}{.}\hspace{.4pt}bbapap\hspace{.1pt}\discretionary{.}{%
}{.}\hspace{.4pt}2014\hspace{.1pt}\discretionary{.}{%
}{.}\hspace{.4pt}08\hspace{.1pt}\discretionary{.}{%
}{.}\hspace{.4pt}015}


\bibitem{Jain99}
A.~K.~o. Jain.
\newblock Data clustering: a review.
\newblock {\em ACM Comput. Surv.}, 31(3):264–323, Sept. 1999. doi: \textsf{%
10\hspace{.1pt}\discretionary{.}{%
}{.}\hspace{.4pt}1145\discretionary{/}{%
}{/}331499\hspace{.1pt}\discretionary{.}{%
}{.}\hspace{.4pt}331504}


\bibitem{jeon22vis}
Jeon et~al.
\newblock Uniform manifold approximation with two-phase optimization.
\newblock In {\em 2022 IEEE Visualization and Visual Analytics (VIS)}, pp. 80--84, 2022. doi: \textsf{%
10\hspace{.1pt}\discretionary{.}{%
}{.}\hspace{.4pt}1109\discretionary{/}{%
}{/}VIS54862\hspace{.1pt}\discretionary{.}{%
}{.}\hspace{.4pt}2022\hspace{.1pt}\discretionary{.}{%
}{.}\hspace{.4pt}00025}


\bibitem{jeon23vis}
H.~Jeon et~al.
\newblock Zadu: A python library for evaluating the reliability of dimensionality reduction embeddings.
\newblock In {\em 2023 IEEE Visualization and Visual Analytics (VIS)}, pp. 196--200, 2023. doi: \textsf{%
10\hspace{.1pt}\discretionary{.}{%
}{.}\hspace{.4pt}1109\discretionary{/}{%
}{/}VIS54172\hspace{.1pt}\discretionary{.}{%
}{.}\hspace{.4pt}2023\hspace{.1pt}\discretionary{.}{%
}{.}\hspace{.4pt}00048}


\bibitem{jeon25tpami}
H.~Jeon et~al.
\newblock Measuring the validity of clustering validation datasets.
\newblock {\em IEEE Transactions on Pattern Analysis and Machine Intelligence}, pp. 1--14, 2025. doi: \textsf{%
10\hspace{.1pt}\discretionary{.}{%
}{.}\hspace{.4pt}1109\discretionary{/}{%
}{/}TPAMI\hspace{.1pt}\discretionary{.}{%
}{.}\hspace{.4pt}2025\hspace{.1pt}\discretionary{.}{%
}{.}\hspace{.4pt}3548011}


\bibitem{jeon25chi}
H.~Jeon, H.~Lee, Y.-H. Kuo, T.~Yang, D.~Archambault, S.~Ko, T.~Fujiwara, K.-L. Ma, and J.~Seo.
\newblock Unveiling high-dimensional backstage: A survey for reliable visual analytics with dimensionality reduction.
\newblock In {\em Proceedings of the 2025 CHI Conference on Human Factors in Computing Systems}, CHI '25. Association for Computing Machinery, New York, NY, USA, 2025. doi: \textsf{%
10\hspace{.1pt}\discretionary{.}{%
}{.}\hspace{.4pt}1145\discretionary{/}{%
}{/}3706598\hspace{.1pt}\discretionary{.}{%
}{.}\hspace{.4pt}3713551}


\bibitem{jeon2025arxiv}
H.~Jeon, J.~Park, S.~Shin, and J.~Seo.
\newblock Stop misusing t-sne and umap for visual analytics, 2025. doi: \textsf{%
10\hspace{.1pt}\discretionary{.}{%
}{.}\hspace{.4pt}48550\discretionary{/}{%
}{/}arXiv\hspace{.1pt}\discretionary{.}{%
}{.}\hspace{.4pt}2506\hspace{.1pt}\discretionary{.}{%
}{.}\hspace{.4pt}08725}


\bibitem{joia11}
P.~Joia, D.~Coimbra, J.~A. Cuminato, F.~V. Paulovich, and L.~G. Nonato.
\newblock Local affine multidimensional projection.
\newblock {\em IEEE Transactions on Visualization and Computer Graphics}, 17(12):2563--2571, 2011. doi: \textsf{%
10\hspace{.1pt}\discretionary{.}{%
}{.}\hspace{.4pt}1109\discretionary{/}{%
}{/}TVCG\hspace{.1pt}\discretionary{.}{%
}{.}\hspace{.4pt}2011\hspace{.1pt}\discretionary{.}{%
}{.}\hspace{.4pt}220}


\bibitem{kruskal64}
J.~B. Kruskal.
\newblock Multidimensional scaling by optimizing goodness of fit to a nonmetric hypothesis.
\newblock {\em Psychometrika}, 29(1):1–27, 1964. doi: \textsf{%
10\hspace{.1pt}\discretionary{.}{%
}{.}\hspace{.4pt}1007\discretionary{/}{%
}{/}BF02289565}


\bibitem{kruskal64psy}
J.~B. Kruskal.
\newblock Nonmetric multidimensional scaling: A numerical method.
\newblock {\em Psychometrika}, 29(2):115–129, 1964. doi: \textsf{%
10\hspace{.1pt}\discretionary{.}{%
}{.}\hspace{.4pt}1007\discretionary{/}{%
}{/}BF02289694}


\bibitem{laber2024}
E.~S. Laber and M.~Bastista.
\newblock On the cohesion and separability of average-link for hierarchical agglomerative clustering, 2024. doi: \textsf{%
10\hspace{.1pt}\discretionary{.}{%
}{.}\hspace{.4pt}48550\discretionary{/}{%
}{/}arXiv\hspace{.1pt}\discretionary{.}{%
}{.}\hspace{.4pt}2411\hspace{.1pt}\discretionary{.}{%
}{.}\hspace{.4pt}05097}


\bibitem{Lavie07}
A.~Lavie and A.~Agarwal.
\newblock Meteor: an automatic metric for mt evaluation with high levels of correlation with human judgments.
\newblock In {\em Proceedings of the Second Workshop on Statistical Machine Translation}, StatMT '07, p. 228–231, 2007.

\bibitem{john09}
J.~A. Lee and M.~Verleysen.
\newblock Quality assessment of dimensionality reduction: Rank-based criteria.
\newblock {\em Neurocomputing}, 72(7):1431--1443, 2009.
\newblock Advances in Machine Learning and Computational Intelligence. doi: \textsf{%
10\hspace{.1pt}\discretionary{.}{%
}{.}\hspace{.4pt}1016\discretionary{/}{%
}{/}j\hspace{.1pt}\discretionary{.}{%
}{.}\hspace{.4pt}neucom\hspace{.1pt}\discretionary{.}{%
}{.}\hspace{.4pt}2008\hspace{.1pt}\discretionary{.}{%
}{.}\hspace{.4pt}12\hspace{.1pt}\discretionary{.}{%
}{.}\hspace{.4pt}017}


\bibitem{lin04}
C.-Y. Lin.
\newblock {ROUGE}: A package for automatic evaluation of summaries.
\newblock In {\em Text Summarization Branches Out}, pp. 74--81. Association for Computational Linguistics, Barcelona, Spain, July 2004.

\bibitem{machado25}
A.~Machado, M.~Behrisch, and A.~Telea.
\newblock Necessary but not sufficient: Limitations of projection quality metrics.
\newblock {\em Computer Graphics Forum}, 2025. doi: \textsf{%
10\hspace{.1pt}\discretionary{.}{%
}{.}\hspace{.4pt}1111\discretionary{/}{%
}{/}cgf\hspace{.1pt}\discretionary{.}{%
}{.}\hspace{.4pt}70101}


\bibitem{mathur2020}
N.~Mathur, T.~Baldwin, and T.~Cohn.
\newblock Tangled up in bleu: Reevaluating the evaluation of automatic machine translation evaluation metrics, 01 2020. doi: \textsf{%
10\hspace{.1pt}\discretionary{.}{%
}{.}\hspace{.4pt}18653\discretionary{/}{%
}{/}v1\discretionary{/}{%
}{/}2020\hspace{.1pt}\discretionary{.}{%
}{.}\hspace{.4pt}acl\discretionary{%
}{-}{-}main\hspace{.1pt}\discretionary{.}{%
}{.}\hspace{.4pt}448}


\bibitem{moor20iclm}
Moor et~al.
\newblock Topological autoencoders.
\newblock In {\em Proceedings of the 37th International Conference on Machine Learning}, vol. 119 of {\em Proceedings of Machine Learning Research}, pp. 7045--7054, 2020.

\bibitem{Moseley23}
B.~Moseley and J.~R. Wang.
\newblock Approximation bounds for hierarchical clustering: average linkage, bisecting k-means, and local search.
\newblock {\em J. Mach. Learn. Res.}, 24(1), Jan. 2023.

\bibitem{Murtagh11}
F.~Murtagh and P.~Contreras.
\newblock Algorithms for hierarchical clustering: an overview.
\newblock {\em WIREs Data Mining and Knowledge Discovery}, 2:86--97, 2011. doi: \textsf{%
10\hspace{.1pt}\discretionary{.}{%
}{.}\hspace{.4pt}1002\discretionary{/}{%
}{/}widm\hspace{.1pt}\discretionary{.}{%
}{.}\hspace{.4pt}53}


\bibitem{muller2022}
D.~Müller et~al.
\newblock Towards a guideline for evaluation metrics in medical image segmentation, 2022. doi: \textsf{%
10\hspace{.1pt}\discretionary{.}{%
}{.}\hspace{.4pt}48550\discretionary{/}{%
}{/}arXiv\hspace{.1pt}\discretionary{.}{%
}{.}\hspace{.4pt}2202\hspace{.1pt}\discretionary{.}{%
}{.}\hspace{.4pt}05273}


\bibitem{Nielsen2016}
F.~Nielsen.
\newblock {\em Hierarchical Clustering}, pp. 195--211.
\newblock Springer International Publishing, Cham, 2016. doi: \textsf{%
10\hspace{.1pt}\discretionary{.}{%
}{.}\hspace{.4pt}1007\discretionary{/}{%
}{/}978\discretionary{%
}{-}{-}3\discretionary{%
}{-}{-}319\discretionary{%
}{-}{-}21903\discretionary{%
}{-}{-}5\_8}


\bibitem{Papin02}
K.~Papineni, S.~Roukos, T.~Ward, and W.-J. Zhu.
\newblock Bleu: a method for automatic evaluation of machine translation.
\newblock In {\em Proceedings of the 40th Annual Meeting on Association for Computational Linguistics}, p. 311–318, 2002. doi: \textsf{%
10\hspace{.1pt}\discretionary{.}{%
}{.}\hspace{.4pt}3115\discretionary{/}{%
}{/}1073083\hspace{.1pt}\discretionary{.}{%
}{.}\hspace{.4pt}1073135}


\bibitem{Rui16}
L.~Rui, H.~Nejati, and N.-M. Cheung.
\newblock Dimensionality reduction of brain imaging data using graph signal processing.
\newblock In {\em 2016 IEEE International Conference on Image Processing (ICIP)}, pp. 1329--1333, 2016. doi: \textsf{%
10\hspace{.1pt}\discretionary{.}{%
}{.}\hspace{.4pt}1109\discretionary{/}{%
}{/}ICIP\hspace{.1pt}\discretionary{.}{%
}{.}\hspace{.4pt}2016\hspace{.1pt}\discretionary{.}{%
}{.}\hspace{.4pt}7532574}


\bibitem{satopaa11icdcws}
V.~Satopaa, J.~Albrecht, D.~Irwin, and B.~Raghavan.
\newblock Finding a "kneedle" in a haystack: Detecting knee points in system behavior.
\newblock In {\em 2011 31st International Conference on Distributed Computing Systems Workshops}, pp. 166--171, 2011. doi: \textsf{%
10\hspace{.1pt}\discretionary{.}{%
}{.}\hspace{.4pt}1109\discretionary{/}{%
}{/}ICDCSW\hspace{.1pt}\discretionary{.}{%
}{.}\hspace{.4pt}2011\hspace{.1pt}\discretionary{.}{%
}{.}\hspace{.4pt}20}


\bibitem{sips09}
M.~Sips, B.~Neubert, J.~P. Lewis, and P.~Hanrahan.
\newblock {Selecting Good Views of High-dimensional Data using Class Consistency}.
\newblock {\em Computer Graphics Forum}, 2009. doi: \textsf{%
10\hspace{.1pt}\discretionary{.}{%
}{.}\hspace{.4pt}1111\discretionary{/}{%
}{/}j\hspace{.1pt}\discretionary{.}{%
}{.}\hspace{.4pt}1467\discretionary{%
}{-}{-}8659\hspace{.1pt}\discretionary{.}{%
}{.}\hspace{.4pt}2009\hspace{.1pt}\discretionary{.}{%
}{.}\hspace{.4pt}01467\hspace{.1pt}\discretionary{.}{%
}{.}\hspace{.4pt}x}


\bibitem{Spearman1987}
C.~Spearman.
\newblock The proof and measurement of association between two things.
\newblock {\em The American Journal of Psychology}, 100(3/4):441--471, 1987.

\bibitem{Taha2015}
A.~A. Taha and A.~Hanbury.
\newblock Metrics for evaluating 3d medical image segmentation: analysis, selection, and tool.
\newblock {\em BMC Medical Imaging}, 15, 2015. doi: \textsf{%
10\hspace{.1pt}\discretionary{.}{%
}{.}\hspace{.4pt}1186\discretionary{/}{%
}{/}s12880\discretionary{%
}{-}{-}015\discretionary{%
}{-}{-}0068\discretionary{%
}{-}{-}x}


\bibitem{Thrun23}
M.~C. Thrun, J.~Märte, and Q.~Stier.
\newblock Analyzing quality measurements for dimensionality reduction.
\newblock {\em Machine Learning and Knowledge Extraction}, 5(3):1076--1118, 2023. doi: \textsf{%
10\hspace{.1pt}\discretionary{.}{%
}{.}\hspace{.4pt}3390\discretionary{/}{%
}{/}make5030056}


\bibitem{venna06}
J.~Venna and S.~Kaski.
\newblock Local multidimensional scaling.
\newblock {\em Neural Networks}, 19(6):889--899, 2006.
\newblock Advances in Self Organising Maps - WSOM’05. doi: \textsf{%
10\hspace{.1pt}\discretionary{.}{%
}{.}\hspace{.4pt}1016\discretionary{/}{%
}{/}j\hspace{.1pt}\discretionary{.}{%
}{.}\hspace{.4pt}neunet\hspace{.1pt}\discretionary{.}{%
}{.}\hspace{.4pt}2006\hspace{.1pt}\discretionary{.}{%
}{.}\hspace{.4pt}05\hspace{.1pt}\discretionary{.}{%
}{.}\hspace{.4pt}014}


\bibitem{wang2021}
Y.~Wang, H.~Huang, C.~Rudin, and Y.~Shaposhnik.
\newblock Understanding how dimension reduction tools work: An empirical approach to deciphering t-sne, umap, trimap, and pacmap for data visualization, 12 2020. doi: \textsf{%
10\hspace{.1pt}\discretionary{.}{%
}{.}\hspace{.4pt}48550\discretionary{/}{%
}{/}arXiv\hspace{.1pt}\discretionary{.}{%
}{.}\hspace{.4pt}2012\hspace{.1pt}\discretionary{.}{%
}{.}\hspace{.4pt}04456}


\end{thebibliography}
\end{document}